# Fairness in Survival Analysis: A Novel Conditional Mutual Information Augmentation Approach

Tianyang Xie and Yong Ge, University of Arizona, {tianyangxie, yongge}@arizona.edu

## ABSTRACT

Survival analysis, a vital tool for predicting the time to event, has been used in many domains such as healthcare, criminal justice, and finance. Like classification tasks, survival analysis can exhibit bias against disadvantaged groups, often due to biases inherent in data or algorithms. Several studies in both the IS and CS communities have attempted to address fairness in survival analysis. However, existing methods often overlook the importance of prediction fairness at pre-defined evaluation time points, which is crucial in real-world applications where decision making often hinges on specific time frames. To address this critical research gap, we introduce a new fairness concept: equalized odds (EO) in survival analysis, which emphasizes prediction fairness at pre-defined time points. To achieve the EO fairness in survival analysis, we propose a Conditional Mutual Information Augmentation (CMIA) approach, which features a novel fairness regularization term based on conditional mutual information and an innovative censored data augmentation technique. Our CMIA approach can effectively balance prediction accuracy and fairness, and it is applicable to various survival models. We evaluate the CMIA approach against several state-of-the-art methods within three different application domains, and the results demonstrate that CMIA consistently reduces prediction disparity while maintaining good accuracy and significantly outperforms the other competing methods across multiple datasets and survival models (e.g., linear COX, deep AFT).

**Keywords**: survival analysis, equalized odds, fairness, pre-defined evaluation time points, conditional mutual information, censored data augmentation

**Fairness in Survival Analysis: A Novel Conditional Mutual Information Augmentation Approach**

## 1. INTRODUCTION

Survival analysis is a set of statistical methods designed to model data where the outcome of interest is the time to the occurrence of a particular event (P. Wang et al., 2017). It is widely applied across many domains, such as healthcare (Khuri et al., 2005; Reddy et al., 2015), education (Ameri et al., 2016), business intelligence (Li et al., 2016; Rakesh et al., 2016), etc. In these applications, survival analysis provides likelihood estimation for the occurrence of events over time, which is useful for a lot of crucial decision making. For example, in healthcare, survival analysis models predict the risk of patient's death and the prediction help providers with strategizing treatment and distribution of medical resources (Keya et al., 2021; Maryland Department of Health, 2019); in criminal justice, survival models are used to estimate the likelihood of recidivism (Angwin et al., 2022; Mattu, 2016), and the estimation is then used for parole decisions and allocation of social resources; in finance, survival models are applied to assess default risks, and loan and investment decisions are further made based on the risk assessment (Stepanova & Thomas, 2002).

However, like other prediction tasks (e.g., classification), survival analysis can exhibit bias against disadvantaged groups, often due to biases inherent in the data or algorithms (Mehrabi et al., 2021, Keya et al., 2021). Such bias can lead to discriminatory decisions, disproportionately affecting individuals in different groups. For instance, in healthcare, when historical records reflect systematic inequalities — such as unequal access to medical care or differences in treatment based on demographic factors — a survival model trained on such data may yield poorer predictions for minority groups, thereby exacerbating existing inequalities (Barocas & Selbst, 2016; Keya et al., 2021); in criminal justice system, if historical data is biased against certain racial or socioeconomic groups, survival models built with the data might overestimate recidivism risks for disadvantaged groups, leading to unjust decision-making (Angwin et al., 2022). Therefore, it is very important to investigate and address potential bias or disparity in survival analysis across diverse application domains.

There are a few recently developed studies on addressing fairness of survival analysis, which form two groups. The first group defines fairness based on the disparity of predicted outcome (i.e., the predicted likelihood of event occurrence) across different subject groups. To achieve the defined fairness, Keya et al. (2021) proposed a fairness regularization term to minimize the absolute difference of predicted outcome across demographic groups and Do et al. (2023) designed a fairness regularization term based on mutual information. Besides, Rahman & Purushotham (2022) adopted Keya et al.'s (2021) fairness regularization design to achieve similar predicted outcomes between the censored and uncensored groups. The second category of work considers fairness as the disparity of prediction accuracy across subject groups, which takes into account both prediction outcome and ground truth information (e.g., actual event occurrence). For example, Zhang & Weiss (2022) introduced a fairness metric, concordance imparity, defined as the worst-case difference in concordance index (i.e., an accuracy metric for survival analysis (Harrell et al., 1982)) between groups, and they further developed a modified random forest algorithm to improve concordance imparity-based fairness. With the same fairness metric, Hu & Chen (2024) incorporated a distributionally robust optimization (DRO) framework for training survival models to improve survival analysis fairness.

However, both lines of research present limitations in many practical applications of survival analysis. As noted by Hu & Chen (2024), the fairness definition in *the first category* of work is constrained by the assumption that the underlying event occurrence likelihood is the same across different subject groups, which is often unrealistic. For example, in healthcare, studies have found that men have a higher mortality rate than women following a breast cancer diagnosis (F. Wang et al., 2019); when those methods in the first category are applied to achieve fair survival analysis in this scenario, it will misalign predictions with real-world patterns, thus significantly compromising overall predictive performance. While *the second category* of work aims to achieve fair prediction accuracy, these approaches still fall short of practical needs in many domains, as they measure fairness over the entire evaluation time period. For instance, the aforementioned concordance imparity, a specific fairness metric proposed in (Zhang & Weiss 2022), quantifies the difference of prediction accuracy between groups across each possible time point within the evaluation period. In other words, the objective of these methods is to achieve fair prediction accuracy along continuous time (i.e., all time points) within the evaluation period. However, in many application domains, survival analysis is used to predict event occurrence across pre-defined and discrete time points. For example, in healthcare, practitioners use survival analysis to forecast mortality risk over a fixed period (e.g., 30 days), rather than across all time points (Khuri et al., 2005); in crowdfunding, survival models are employed to estimate the likelihood of project success within a pre-defined timeframe such as 60 days (Li et al., 2016); in recidivism prediction, survival analysis is applied to predict the likelihood of recidivism over three years, which aligns with the average duration of a federal supervision term (*Just the Facts*, 2018). Therefore, achieving survival analysis fairness across such pre-defined time points is critical. Unfortunately, fairness metrics (e.g., concordance imparity) proposed in the second category of work do not meet this need because they place undue emphasis on less important time points, instead of those pre-defined ones (e.g., 30 days), and those methods developed based on these metrics cannot yield desired prediction fairness across the pre-defined time intervals.

To address the limitations of prior works, we introduce a new fairness measure, equalized odds (EO), for survival analysis. This metric ensures equitable prediction accuracy between groups across each pre-defined time interval, rather than all time points. Specifically, within each of the pre-defined time intervals, it requires independence between predicted outcome (e.g., recidivism) and sensitive attributes (e.g., gender), conditioned on the true labels (i.e., the indicator of event occurrence). This fairness measure for survival analysis aligns with the equalized odds (EO) in classification fairness (Hardt et al., 2016). But there is a key difference between them: our proposed EO for survival analysis requires the conditional independence across all pre-specified time points, whereas EO in classification does not consider time.

With our proposed fairness measure, we develop a novel approach, namely Conditional Mutual Information Augmentation (CMIA), to enhance the fairness of survival analysis. CMIA approach features two key innovations: a novel fairness regularization term based on conditional mutual information (CMI) statistics (Wyner, 1978), and a new censored data augmentation technique. The fairness regularization term, grounded in information theory, ensures that the survival model adheres to the equalized odds in survival analysis. The censored data augmentation module enhances the balance between accuracy and fairness during model training. Both components are integrated in a joint learning objective. The CMIA approach is model agnostic and thus is adaptable to a variety of survival models. Using multiple datasets and different survival models, we conduct intensive empirical evaluations to demonstrate the superiority of our developed CMIA approach by comparing with several existing methods that indeed fail to achieve the defined EO fairness. We also employ visualizations to interpret how the censored data augmentation helps the method navigate the trade-off between fairness and accuracy (i.e., improved fairness with minimal impact on accuracy).

## 2. RELATED WORKS

In this section, we review the related works on survival models and fairness in survival analysis.

### 2.1. Survival Analysis Models

The extant survival analysis models can be categorized into three groups (P. Wang et al., 2017): non-parametric, parametric, and semi-parametric. Below, we elaborate on each category.

Non-parametric models make no assumptions about the underlying data distribution. The most widely used one is the Kaplan-Meier (KM) estimator (Kaplan & Meier, 1958), which calculates survival probability up to a specific time as a cumulative product of prior conditional survival probabilities. On the other hand, the Nelson-Aalen estimator (NA) (Andersen et al., 2012) employs counting process techniques to estimate survival probabilities. Another well-known method is Life Table (LT) analysis (Cutler & Ederer, 1958), which organizes survival data into tables and provides visual analysis, making it especially useful for grouped data.

Parametric models, on the other hand, assume that the survival time -- the time to the event occurrence -- follows a specific distribution, such as Weibull and Normal distributions. Maximum likelihood estimation (MLE) (Lee, 2003) is used to estimate parameters and predict survival time under these assumptions. When the logarithm of survival time is expressed as a linear function of covariates, the model (i.e., the distribution assumption) is known as an Accelerated Failure Time (AFT) model (Kalbfleisch & Prentice, 2011), which allows for the use of generalized linear regression techniques for efficient estimation.

Finally, semi-parametric models combine elements of both parametric and non-parametric approaches. The cornerstone semi-parametric model is the Cox Proportional Hazards (COX) model (Cox, 1972). Like non-parametric models, it makes no assumption about the distribution of survival times. At the same time, it incorporates a parametric loss function to explain the risk of the event occurrence. Building on the original COX model, (Katzman et al., 2018) extended the linear relationship between covariates and the risk of the event occurrence to a deep model framework, allowing for more complex and nonlinear relationship modeling.

In this study, we consider Cox and AFT models as they have been widely used in many survival analysis applications. In Section 3.1, we will review their technical details, based on which we propose our EO fairness measure and CMIA approach.

## 2.2. Fairness in Survival Analysis

Prior studies on fairness of survival analysis can be grouped into two categories.

*The first strategy* aims to reduce disparities in predicted outcomes, specifically the likelihood of an event occurring, across different groups. As discussed in Section 1, the works of Keya et al. (2021), Do et al. (2023), and Rahman & Purushotham (2022) align with this approach. The key distinction lies in their methods: Keya et al. (2021) proposed a fairness regularization term based on minimizing the absolute differences in predicted outcomes between groups, while Do et al. (2023) designed their regularization using mutual information. In contrast, Rahman & Purushotham (2022) shifted the focus to addressing disparities caused by data censorship rather than demographic differences, adopting Keya et al.'s (2021) design.

*The second strategy* takes a different route by emphasizing fairness in predictive accuracy rather than just predicted outcomes. This approach considers not only the model's predictions but also the actual observed data. For instance, Zhang & Weiss (2022) developed a metric called concordance imparity, which quantifies the largest discrepancy in concordance index, and further designed a modified random forest algorithm to improve fairness according to this metric. Later, Hu & Chen (2024) introduced a distributionally robust optimization (DRO) framework, also aiming to address fairness in predictive accuracy.

However, none of these prior studies have addressed the fairness of survival analysis predictions that are performed at pre-define time points (e.g., every 30 days), which is very needed in many real-world applications such as recidivism prediction. This paper aims to address this critical research gap, and our proposed EO fairness measure for survival analysis and designed CMIA approach are fundamentally different from the prior studies.

## 3. METHODOLOGY

In this section, we introduce our proposed methodology. First, we outline preliminaries, including the problem formulation and survival analysis models. Next, we formally define equalized odds fairness within the context of survival analysis. Following this, we introduce our proposed Conditional Mutual Information Augmentation method, which consists of two core components, i.e., Fairness Regularization and Censor Data Augmentation, and present the formalized learning objective.

### 3.1. Preliminaries

In this subsection, we provide the background on survival analysis relevant to this study. We first formalize the survival analysis problem and then briefly introduce several survival models.

#### 3.1.1. Survival Analysis Problem

In survival analysis, each observation (e.g., a patient's medical record) is represented as a four-tuple $\{(X, T, \delta, Z)\}_{i=1}^{n}$, where $n$ is the number of observations. Here, $X \in \mathbb{R}^{p}$ denotes the feature vector, which may include attributes like age, income, and social status. The binary indicator $\delta \in \{0,1\}$ specifies whether the event of interest (e.g., death) occurs during the study period. $T \in \mathbb{R}$ represents the time duration: when $\delta = 1$, $T$ is the time-to-event, i.e., the time from the study's start to the occurrence of the event; when $\delta = 0$ (i.e., there is no occurrence of event during the study period), T is the time until the study concludes, marking the observation as "censored" which means the event may still happen in the future and its timing remains unknown. Finally, $Z \in C$ is a categorical variable representing a sensitive attribute, such as ethnicity or gender of patient. Given a set of observations, survival analysis problem is to train a model to predict the likelihood that one event occurs after time duration $t$ for each single record.

### 3.1.2. Survival Analysis Models

To illustrate the concept of survival analysis models, we begin with a foundational example: the Cox Proportional Hazards (COX) model (Cox, 1972). We then expand our discussion to other models by modifying key components of the COX model.

The COX model has two fundamental functions: (1) $S(t|X) = \mathrm{P}(T > t|X)$, which is the survival function, denoting the probability of event occurrence after time duration $t$, or in other words, the subject has "survived" in time duration t without the event occurrence. The estimation of survival function is the ultimate target for survival analysis because once we successfully estimate it, we can use it to predict survival probability for any time duration. (2) $\lambda(t|X) = \lim_{\eta \to 0} \frac{\mathrm{P}(t < T < t+\eta | T > t, X)}{\eta}$, which is the hazard function, denoting the probability that an individual will not survive an extra infinitesimal amount of time η, given that she has already survived for time during $t$. These two functions have the following relationship:

$$S(t|X) = \exp\left(-\int_0^t \lambda(t|X)\mathrm{d}t\right). \tag{1}$$

The COX model further assumes that the hazard function can be decomposed into a time dependent component $\lambda_0(t)$ and a feature dependent component $g_\theta(X)$:

$$\lambda(t|X) = \lambda_0(t) \cdot \exp\big(g_\theta(X)\big), \tag{2}$$

where $\theta \in \Theta$ are model parameters. Additionally, in the original COX model, the $g_\theta(X)$ function is assumed to be linear: $g_\theta(X) = X\theta$. The estimation of $g_\theta(X)$ is performed by optimizing:

$$min_{g_\theta} L(g_\theta) = \sum_{i=1}^{n} \delta_i \left[ -g_\theta(X_i) + log \sum_{j \in R_i} exp\big(g_\theta(X_i)\big) \right]. \tag{3}$$

Afterward, $\lambda_0(t)$ can be estimated by a non-parametric method, such as the Kaplan-Meier (KM) estimator (Kaplan & Meier, 1958), allowing $S(t|X)$ to be calculated using Eq. (1).

The form of $g_\theta(X)$ is flexible. For example, Katzman et al. (2018) explored deep COX models by redefining $g_\theta(X)$ as: $g_\theta(X) = MLP_\theta(X)$, where $MLP_\theta$ represents a Multi-layer Perceptron (MLP) (Goodfellow et al., 2016) function. The training and prediction workflow remains unchanged. Furthermore, the objective function in Eq. (3) can also be modified. For example, when the objective function is redefined as:

$$min_{g_\theta,\sigma} L(g_\theta) = \sum_{i=1}^{n} [\delta_i(\log(\sigma) - w_i) + \exp(w_i)], \tag{4}$$

where $w_i = \frac{\log(T_i) - g_\theta(X_i)}{\sigma}$, the formulation becomes the well-known AFT model, where $\sigma \in \mathbb{R}^+$ is an additional parameter that is jointly learned alongside the model parameters $\theta$. Accordingly, the survival function in the AFT model is updated as:

$$S(t|X) = \exp\left(-\left(\frac{t}{\exp\big(g_\theta(X_i)\big)}\right)^{\frac{1}{\sigma}}\right) \tag{5}$$

This difference between both formulated objective functions (i.e., Eq. (3), (4)) indeed reflects a different assumption about the underlying data pattern.

In this study, we consider four survival model scenarios, formed by combining two different formulations of $g_\theta(X)$ (i.e., Linear vs Deep models) with two different learning objective functions (COX vs AFT objectives). As our proposed CMIA approach is model agnostic, we will marry it with each of the four survival model scenarios and evaluate its performance. Next, we introduce our proposed new fairness measure, equalized odds (EO), for survival analysis.

## 3.2. Equalized Odds Fairness in Survival Analysis

As introduced in Section 1, our proposed EO fairness requires conditionally independence between predicted outcome and a sensitive attribute, and it aims to achieve equitable prediction accuracy between groups across each pre-defined time interval (e.g., 30 days). Mathematically, the EO fairness requires a survival model to satisfy the following conditional independence:

$$\widehat{Y}_t \perp Z | Y_t, \quad \forall t \in Q , \tag{6}$$

where $Y_t := I(T \leq t, \ \delta = 1)$ is a binary variable indicating whether the event occurs before time $t$ for a given observation; $\widehat{Y}_t$ represents the predicted binary outcome inferred by the survival function $S(t|X)$. $\widehat{Y}_t$ is usually assigned a value of 1 when $S(t|X)$ is less than 0.5, and otherwise it is assigned a value of 0; The set $Q$ refers to a collection of pre-specified evaluation time points, which is often determined based on the practical needs in different applications.

Like EO in classification, EO for survival analysis has the following *necessary and sufficient condition*: the equality of true positive rate (TPR) and false positive rate (FPR) across all sensitive-attribute groups: $TPR_{c_1,t} = \cdots = TPR_{c_{|C|},t}; FPR_{c_1,t} = \cdots = FPR_{c_{|C|},t}, \forall t \in Q$, where $TPR_{c_i,t} = P(\widehat{Y}_t = 1 | Y_t = 1, Z = c_i)$ and $FPR_{c_i,t} = P(\widehat{Y}_t = 1 | Y_t = 0, Z = c_i)$, respectively. $|C|$ denotes the number of sensitive-attribute groups. The proof of this condition is provided in Appendix A. Satisfying this condition essentially implies that the survival analysis yields equitable predictive accuracy (quantified by TPR and FPR) between different sensitive-attribute groups across all the pre-specified evaluation time points. This *necessary and sufficient condition* will be later used for evaluating the EO fairness of survival models in Section 4.3.

We would like to remark that unlike the existing two categories of fairness measures reviewed in Section 2.2, our proposed EO fairness aims to achieve fair prediction of event occurrence across the pre-defined time points (e.g., 30 days), which is very needed in many real-world applications (e.g., recidivism event prediction) of survival analysis. Based on this proposed EO fairness, we develop our novel CMIA approach as follows.

### 3.3. Conditional Mutual Information Augmentation (CMIA) Approach

Our proposed CMIA approach consists of two core components: a novel fairness regularization term and a new censored data augmentation module. The fairness regularization term is designed to ensure the survival model meets the EO fairness criteria, and the censored data augmentation module is developed to further enhance the balance between accuracy and fairness of survival analysis prediction. Both components are fused through a joint-learning objective function.

#### 3.3.1. Fairness Regularization

The fairness regularization term, which will be added to the original learning objectives (i.e., Eq. (3) and (4)), is designed to regulate the survival model to meet the EO fairness criterion. Therefore, the fairness regularization term needs to capture how well the survival model satisfies the EO fairness criterion. More specifically, it should be formalized to quantify the degree of conditional independence $\widehat{Y}_t \perp Z | Y_t, \forall t \in Q$ (i.e., Eq. (6)).

To achieve this, we adapt the conditional mutual information (CMI) statistics (Wyner, 1978) from information theory to develop our fairness regularization term. It measures the shared information between two variables, conditioned on a third one. In our context, it is defined as:

$$CMI_t := \sum_{\widehat{Y}_t \in \{0,1\}} \sum_{Y_t \in \{0,1\}} \sum_{Z \in C} P_{\widehat{Y}_t, Z, Y_t} \log \frac{P_{\widehat{Y}_t | Z, Y_t}}{P_{\widehat{Y}_t | Y_t}}, \quad \forall t \in Q \tag{7}$$

where $P_{\widehat{Y}_t, Z, Y_t}$ is the joint probability mass function of $\widehat{Y}_t, Z, Y_t$; $P_{\widehat{Y}_t | Z, Y_t}$ is the probability mass function of $\widehat{Y}_t$ conditional on $Z, Y_t$; $P_{\widehat{Y}_t | Y_t}$ is the probability mass function of $\widehat{Y}_t$ conditional on $Y_t$. With this definition, we identify and prove an important mathematical property of the CMI:

$$CMI_t \geq 0, \quad \text{and} \quad CMI_t = 0 \Longrightarrow \widehat{Y}_t \perp Z | Y_t . \tag{8}$$

In Appendix B, we provide detailed proof of this property. This property informs that ensuring $\text{CMI}_t = 0, \forall t \in Q$ guarantees that the survival model meets the EO fairness criterion. Given its non-negativity, the CMI statistic makes itself an ideal candidate for the fairness regularization term because minimizing the learning objective (which will include the regularization term) could force $\text{CMI}_t(\forall t \in Q)$ to be close to zero.

However, $\text{CMI}_t$ is not directly computable because the probability mass function $P_{\hat{Y}_t, Z, Y_t}$, $P_{\hat{Y}_t | Z, Y_t}$ and $P_{\hat{Y}_t | Y_t}$ are unobserved. This challenge hinders the direct use of the CMI statistics as a fairness regularization term. To address this challenge, we propose a novel and computable approximation method to approximate the CMI statistics. Specifically, it is defined as:

$$\widehat{CMI}_{m,t} := \frac{1}{n}\sum_{i=1}^{n}\frac{1}{m}\sum_{j=1}^{m} log \left[ \frac{\frac{1}{n_{Y_{i,t}Z_i}}\sum_{Y_{l,t}=Y_{i,t},Z_l=Z_i}\phi_\tau\left(\epsilon_j + g_\theta(X_i) - g_\theta(X_l)\right)}{\frac{1}{n_{Y_{i,t}}}\sum_{Y_{k,t}=Y_{i,t}}\phi_\tau\left(\epsilon_j + g_\theta(X_i) - g_\theta(X_k)\right)} \right], \quad \forall t \in Q \quad (9)$$

where $m$ is a pre-determined large number; $n_{Y_{i,t}Z_i}$ denotes the number of observations that has the same $Y_t$ and $Z$ values as the observation $i$ at time $t$; Similarly, $n_{Y_{i,t}}$ denotes the number of observations that has the same $Y_t$ value as the observation $i$ at time $t$; $\phi_\tau$ is the probability density function of a normal distribution $N(0, \tau)$ and $\tau$ is a pre-determined hyperparameter; $\epsilon_j$ is a random noise drawn from this normal distribution; the computation of $\widehat{CMI}_{m,t}$ involves drawing the random noise $\epsilon$ repeatedly $m$ times from the normal distribution $N(0, \tau)$. With these mathematical formalizations, we identify that this approximation (i.e., $\widehat{CMI}_{m,t}$) possesses similar properties as the original $CMI$ does:

$$\widehat{CMI}_{m,t} \geq 0, \text{and } \widehat{CMI}_{m,t} = 0 \Longrightarrow \hat{Y}_t \perp Z | Y_t, \quad \text{if } m \longrightarrow \infty. \quad (10)$$

In Appendix C, we provide detailed proof for this property. The property implies that when the number of drawn random noise approaches infinity, the approximation term $\widehat{CMI}_{m,t}$ becomes non-negative and minimizing it to zero will lead to the satisfaction of the EO fairness criterion. As $\widehat{\text{CMI}}_{m,t}$ is computable, we finalize the proposed fairness regularization term as:

$$R_{EO}(g_\theta) := \sum_{t \in Q} \widehat{CMI}_{m,t} \,. \tag{11}$$

This term will be added to the original learning objectives (i.e., Eq. (3) and (4)). Then by minimizing the learning objective, $R_{EO}$ is forced to approach zero, compelling the learned survival model to meet the EO fairness criteria.

### 3.3.2. Censored Data Augmentation

While our developed fairness regularization term imposes the survival model to satisfy the defined EO fairness, it remains a challenge how to strike a balance between accuracy and fairness in survival analysis. Such a contradiction between accuracy and fairness exists in many fairness-aware prediction tasks because optimizing fairness often does not align with optimization of accuracy during the learning process (Fu et al., 2021; Sener & Koltun, 2018). To mitigate this challenge, we propose a novel censored data augmentation approach to improve the balance between accuracy and fairness. Intuitively, the augmented data serves as a "mediation zone," where the objectives of optimizing accuracy and fairness find common ground. In this zone, both objectives tend to compromise slightly throughout the optimization process, allowing the parameter estimation to eventually achieve a state that benefits both perspectives.

Our data augmentation approach draws inspiration from the uniqueness of survival analysis data, which is that some event occurrences in the observations are "censored." In other words, for observations with $\delta = 0$, the time of event occurrence is unknown. During survival model training (via minimizing the learning objective), these censored observations contribute by signaling that the event has not occurred within the study period, thus encouraging the survival model to assign lower likelihood of event occurrence to similar data instances; compared with uncensored observations that offer precise time information on event occurrence, these censored instances are less informative in guiding the training process because the time to event is missing. Such missing information indeed provides us with a unique opportunity to balance prediction accuracy and fairness through data imputation (Shorten & Khoshgoftaar, 2019), which could alleviate the intricate data bias issue that inherently causes the prediction unfairness (Mehrabi et al., 2021). In fact, extant literature has explored such data imputation approach for improving the balance between recommendation accuracy and fairness (Rastegarpanah et al., 2019), where the imputed data are unobserved user-item ratings. In the context of survival analysis, we propose to impute the time of event occurrence for the censored observations.

For each censored observation $j$, we design the data augmentation as follows. Let $\Delta_j$ denote an additional time duration corresponding to the censored observation $j$. When $\Delta_j > 0$, the module synthesizes an event time $\tilde{T}_j = T_j + \Delta_j$ for the censored observation $j$ and updates the event occurrence indicator as $\tilde{\delta}_j = 1$, where $T_j$ is its original time duration before imputation. When $\Delta_j = 0$, there will be no imputation and the observation $j$ remains unchanged. Assuming there are $n_{\delta=0}$ censored observations in the original dataset, we use $\Delta \in \mathbb{R}^{+^{n_{\delta=0}}}$, a non-negative vector, to denote the *to-be-optimized* additional time durations for all the censored observations. After the data imputation, the augmented dataset can be denoted as:

$$\{(X_i, T_i, \delta_i, Z_i)\}_{\delta_i=1} \cup \left\{\left(X_j, \tilde{T}_j, \tilde{\delta}_j, Z_j\right)\right\}_{\delta_j=0} \tag{12}$$

where $\tilde{T}_j = T_j + \Delta_j$ and $\tilde{\delta}_j = I(\Delta_j > 0)$. In other words, the augmented dataset consists of two parts: the original uncensored records and the possibly updated ones from the original censored observations. For the second part, both the time to event $\tilde{T}_j$ and the event occurrence indicator $\tilde{\delta}_j$ for each original censored observation are determined by the data augmentation.

With the above design, the next challenge is how to optimize the values of $\Delta$, which essentially determines the augmented data. To solve this challenge, we include $\Delta$ as a set of parameters into the learning objective and the values of $\Delta$ will be jointly learned alongside the survival model parameters $\theta$.

### 3.3.3. Joint Learning Objective

The joint learning objective of our CMIA approach is finally formalized as follows:

$$min_{g_\theta,\Delta} L(g_\theta, \Delta) + \lambda_1 R_{EO}(g_\theta, \Delta) + \lambda_2 \lVert\theta\rVert_2^2, \tag{13}$$

where $L(g_\theta, \Delta)$ is the foundational objective function defined in Section 3.1 and $R_{EO}(g_\theta, \Delta)$ is the regularization term introduced in Section 3.3.1. As introduced in Section 3.3.1, minimizing $R_{EO}(g_\theta, \Delta)$ through the learning process will compel the survival model to meet our defined EO fairness. Both terms (i.e., $L(g_\theta, \Delta)$ and $R_{EO}(g_\theta, \Delta)$) have an additional input $\Delta$ because the imputed data (i.e., synthetic uncensored observations) depend on the values of $\Delta$. $\lambda_1$ and $\lambda_2$ are the scale parameters for fairness regularization and the $L_2$-penalty. As the formalization suggests, $\Delta$ will be jointly learned with the survival model parameters $\theta$ by minimizing this objective. Optimizing the joint objective with $\Delta$ as additional parameters enables the survival model to strike a good balance between prediction accuracy and fairness. The optimization of the joint learning objective proceeds similarly to the standard process for training other machine learning models. Specifically, we compute the gradient of the joint learning objective with respect to $\theta$ and $\Delta$, and apply the Adam algorithm (Kingma & Ba, 2017) to perform the optimization.

## 4. EVALUATION

In this section, we evaluate our proposed CMIA approach in comparison to state-of-the-art (SOTA) benchmark methods with three datasets collected in different domains. After introducing the datasets, we present the experiment settings, the SOTA methods, and evaluation metrics. We then present the main results and discuss the findings. Finally, we use visualization to illustrate how the data augmentation improves the balance between accuracy and fairness.

### 4.1. Data

We use prominent datasets collected from healthcare, social justice, and consumer loans to evaluate our CMIA approach. The details of each dataset are outlined below.

**FLC Dataset**: This widely used healthcare dataset was collected from a study that evaluated the ability of the serum immunoglobulin free light chain (FLC) assay to predict overall survival related to immune dysregulation (Dispenzieri et al., 2012; Kyle et al., 2006). It includes 6,524 patients, each described by six features. The goal is to predict the risk of death. During the study, 30% of the patients passed away, while the remaining 70% were marked as censored. The sensitive attribute in this dataset is gender. The minimum, maximum, and average values of the time duration (i.e., T) are 1, 5166, and 3583 days. Accordingly, we predict the risk of death across the time points of 2190, 2920, 3650, and 4380 days.

**Consumer Loan Dataset**: This is a well-known credit scoring dataset used to analyze default rate of consumer loans in the German market (Hofmann, 1994). We use the version suggested by the PySurvival package (Fotso, 2019) to study the speed of loan repayment. The objective of the survival models is to predict the likelihood of full loan repayment during the study period. The dataset contains 1,000 loans, each described by 18 demographic features. In this dataset, 70% of the loans are fully repaid, while the remaining 30% are marked as censored. The sensitive attribute is also gender. The minimum, maximum, and average values of the time duration are 4, 72, and 21 days. We predict the outcome (i.e., full repayment) at 21 and 35 days.

**COMPAS &COMPAS Multi Dataset**: The COMPAS dataset pertains to a system used to predict criminal recidivism, which has faced criticism for potential bias (Angwin et al., 2022). It is primarily employed in bail and sentencing decisions but could also be used to allocate social work resources. The dataset includes 10,314 offenders and six demographic attributes. During the study, 27% of subjects reoffended, with a median event time of 173 days, while the remaining 73% are marked as censored. The sensitive attributes are ethnicity and gender. For simplicity, ethnicity is discretized into a binary variable: Caucasian and Non-Caucasian.

We consider two versions of the COMPAS dataset: COMPAS and COMPAS Multi. The COMPAS version uses binary ethnicity (Caucasian or Non-Caucasian) as the sensitive attribute, while COMPAS Multi includes both ethnicity and gender as sensitive attributes. This results in four groups, which allows us to evaluate our approach on a dataset with multiple sensitive attributes. To align with the COMPAS system's definition of recidivism — "a new misdemeanor or felony offense within two years of the COMPAS administration date" (Mattu, 2016)— we apply the survival models to predict recidivism at 730 days (2 years). Additionally, we consider the time point of 1095 days (3 years), which corresponds to the average length of a federal supervision term (*Just the Facts*, 2018). Therefore, the survival models are used to predict the outcome at two time points, 730 and 1095 days, for both the COMPAS and COMPAS Multi.

### 4.2. Experiment Settings and Benchmark Methods

Each dataset is then randomly split into training, validation, and test sets with ratios of 0.8, 0.1, and 0.1, respectively. Survival models are trained on the training set, hyperparameters are fine-tuned on the validation set, and performance is evaluated on the test set. We do not show the detailed hyperparameter selection here due to space constraints. For each dataset, as outlined in Section 3.1, we consider four survival model scenarios, which are combinations of two $g_\theta(X)$ functions—Linear and Deep—and two foundational objective functions: AFT and COX. In the four scenarios, we evaluate our proposed CMIA approach against three different benchmark methods. Below, we briefly introduce each benchmark method.

**g-Difference (GD) (Keya et al., 2021)**: This method introduces a regularization term based on the absolute difference of groupwise $g_\theta(X)$ function values to achieve group fairness in terms of statistical parity. While the original implementation (Keya et al., 2021) focuses on the COX objective function, we generalize the idea to the AFT objective function as well.

**Distributionally Robust Optimization (DRO) (Hu & Chen, 2024)**: This benchmark method uses distributionally robust optimization to address survival analysis fairness by ensuring equal accuracy across groups defined by the sensitive attribute. Specifically, it minimizes the worst-case error across all subpopulations that exceed a predetermined threshold. As in the original paper, DRO is only applied to the COX objective function.

**Vanilla**: In addition, for each of the four scenarios, we consider the vanilla survival model (e.g., linear $g_\theta(X)$ with COX objective) as an additional benchmark method.

### 4.3. Evaluation Metrics

We employ two widely used metrics to evaluate the accuracy of survival models: the averaged area under the curve (aAUC) and the averaged Brier score (aBrier). The aAUC is defined as: $aAUC = \frac{1}{|Q|}\sum_{t \in Q} AUC_t$, which represents the average AUC value across all evaluation time points. The averaged Brier score is defined as:

$$aBrier = \frac{1}{|Q|}\sum_{t \in Q} Brier_t = \sum_{t \in Q}\left[\frac{1}{n}\sum_{i=1}^{n}\left(I(T_i > t) - P(\widehat{T}_i \leq t)\right)^2\right], \tag{14}$$

which averages the Brier score (Brier, 1950) over all evaluation time points. $P(\widehat{T}_i \leq t)$ represents the predicted probability of the event occurrence before time t. While a higher value of aAUC indicates better performance, a lower value of aBrier reflects higher accuracy.

To evaluate the EO fairness of survival models, we use the averaged maximum difference of the true positive rate (adTPR) and the false positive rate (adFPR), which are defined as:

$$adTPR = \frac{1}{|Q|} \sum_{t \in Q} dTPR_t = \frac{1}{|Q|} \sum_{t \in Q} \left[ max_{c \in C} TPR_{c,t} - min_{c \in C} TPR_{c,t} \right], \qquad (15)$$

$$adFPR = \frac{1}{|Q|} \sum_{t \in Q} dFPR_t = \frac{1}{|Q|} \sum_{t \in Q} \left[ max_{c \in C} FPR_{c,t} - min_{c \in C} FPR_{c,t} \right], \qquad (16)$$

For both metrics, it first computes the max difference of TPR or FPR among different groups (defined by the sensitive attribute) at each time point, and then computes the average of such max difference across all time points. Smaller adTPR and adFPR indicates better EO fairness.

### 4.4. Main Results

We present the main evaluation results in Tables 1-4, from which we have the following observations. First, the vanilla survival models consistently show notable unfairness in terms of Equalized Odds across all datasets and scenarios. For instance, on the FLC dataset, the vanilla models result in an adTPR exceeding 0.0241 and an adFPR higher than 0.0119 across all survival model scenarios. Second, while both GD and DRO cannot consistently mitigate this disparity, our proposed CMIA approach significantly reduces unfairness across all datasets and survival model scenarios. Notably, on the COMPAS dataset, CMIA achieves a mitigation of up to 93.78% in adTPR and 92.75% in adFPR for the deep survival model trained with the Cox objective function. Third, CMIA successfully balances accuracy and fairness. On the FLC dataset, it even improves accuracy while reducing unfairness. For instance, in the scenario of the linear $g_\theta(X)$ trained with the AFT objective function, CMIA improves the aAUC by 4.83% and reduces the aBrier, adTPR, and adFPR by 13.48%, 41.25%, and 60.96%, respectively.

| FLC | | Linear | | | | Deep | | | |
|---|---|---|---|---|---|---|---|---|---|
| | | aAUC↑ | aBrier↓ | adTPR↓ | adFPR↓ | aAUC↑ | aBrier↓ | adTPR↓ | adFPR↓ |
| AFT | Vanilla | 0.6908 | 0.2199 | 0.0241 | 0.0767 | 0.7428 | 0.1946 | 0.0521 | 0.0119 |
| | GD | 0.6929 | 0.2203 | 0.0157 | 0.0611 | **0.7464** | 0.1957 | 0.0313 | 0.0179 |
| | CMIA | **0.7242** | **0.1903** | **0.0142** | **0.0299** | 0.6912 | **0.1866** | **0.0007** | **0.0088** |
| | % | 4.83% | -13.48% | -41.25% | -60.96% | -6.96% | -4.11% | -98.71% | -26.39% |
| COX | Vanilla | 0.7370 | 0.1913 | 0.0499 | 0.0228 | 0.7460 | 0.1889 | 0.0450 | 0.0134 |
| | GD | 0.7352 | 0.1905 | 0.0365 | 0.0256 | **0.7472** | 0.1880 | 0.0362 | **0.0115** |
| | DRO | 0.6031 | 0.2141 | 0.0138 | 0.0195 | 0.5498 | 0.2502 | 0.0519 | 0.0460 |
| | CMIA | **0.7439** | **0.1894** | **0.0101** | **0.0081** | 0.7245 | **0.1862** | **0.0085** | 0.0124 |
| | % | 0.93% | -1.00% | -79.73% | -64.44% | -2.88% | -1.47% | -81.09% | -7.64% |

| Credit | | Linear | | | | Deep | | | |
|---|---|---|---|---|---|---|---|---|---|
| | | aAUC↑ | aBrier↓ | adTPR↓ | adFPR↓ | aAUC↑ | aBrier↓ | adTPR↓ | adFPR↓ |
| AFT | Vanilla | 0.8635 | 0.1369 | 0.0526 | 0.0318 | **0.8656** | 0.1487 | 0.0494 | 0.0476 |
| | GD | 0.8586 | 0.1393 | 0.0290 | 0.0275 | 0.8655 | 0.1342 | 0.0674 | **0.0138** |
| | CMIA | **0.8865** | **0.1251** | **0.0218** | **0.0141** | 0.8576 | **0.1340** | **0.0339** | 0.0284 |
| | % | 2.66% | -8.68% | -58.51% | -55.47% | -0.93% | -9.92% | -31.37% | -40.38% |
| COX | Vanilla | 0.8539 | **0.5553** | 0.0047 | 0.0075 | 0.7499 | **0.3963** | 0.0605 | 0.0672 |
| | GD | **0.8540** | 0.5784 | 0.0081 | 0.0024 | 0.8343 | 0.5654 | 0.0053 | 0.0046 |
| | DRO | 0.7476 | 0.6127 | **0.0010** | 0.0042 | 0.6037 | 0.6194 | **0.0019** | 0.0010 |
| | CMIA | 0.8200 | 0.5953 | 0.0029 | 0.0014 | **0.8445** | 0.5863 | 0.0035 | **0.0000** |
| | % | -3.97% | 7.19% | -38.63% | -82.02% | 12.61% | 47.94% | -94.30% | -99.94% |

| Compas | | Linear | | | | Deep | | | |
|---|---|---|---|---|---|---|---|---|---|
| | | aAUC↑ | aBrier↓ | adTPR↓ | adFPR↓ | aAUC↑ | aBrier↓ | adTPR↓ | adFPR↓ |
| AFT | Vanilla | **0.6535** | 0.3491 | 0.0292 | 0.0769 | 0.6790 | 0.3610 | 0.0351 | 0.0808 |
| | GD | 0.6333 | 0.3573 | 0.0281 | 0.0654 | 0.6759 | 0.3604 | **0.0217** | **0.0270** |
| | CMIA | 0.6444 | **0.1317** | **0.0136** | 0.0376 | **0.6830** | **0.1246** | 0.0262 | 0.0294 |
| | % | -1.40% | -62.29% | -53.26% | -51.04% | 0.59% | -65.49% | -25.41% | -63.61% |
| COX | Vanilla | 0.7262 | 0.4018 | 0.0201 | 0.0764 | **0.7303** | 0.4029 | 0.0278 | 0.0758 |
| | GD | 0.7283 | 0.4016 | 0.0225 | 0.0755 | 0.7290 | **0.3968** | 0.0297 | 0.0590 |
| | DRO | **0.7425** | 0.4028 | **0.0046** | **0.0086** | 0.7027 | 0.4049 | 0.0031 | 0.0136 |
| | CMIA | 0.7225 | **0.3966** | 0.0094 | 0.0181 | 0.6914 | 0.4080 | **0.0017** | **0.0055** |
| | % | -0.51% | -1.28% | -53.14% | -76.30% | -5.33% | 1.26% | -93.78% | -92.75% |

| Compas Multi | | Linear | | | | Deep | | | |
|---|---|---|---|---|---|---|---|---|---|
| | | aAUC↑ | aBrier↓ | adTPR↓ | adFPR↓ | aAUC↑ | aBrier↓ | adTPR↓ | adFPR↓ |
| AFT | Vanilla | 0.6991 | **0.3397** | 0.0857 | 0.1009 | **0.6936** | **0.3430** | 0.0572 | 0.1129 |
| | GD | **0.6903** | 0.2067 | 0.1486 | 0.1133 | 0.6452 | 0.3644 | **0.0205** | **0.0128** |
| | CMIA | 0.6698 | 0.1310 | **0.0424** | **0.0508** | 0.6659 | 0.1215 | 0.0272 | 0.0269 |
| | % | -4.18% | -61.43% | -50.56% | -49.65% | -4.00% | -64.57% | -52.46% | -76.20% |
| COX | Vanilla | **0.7436** | **0.3912** | 0.0718 | 0.1012 | **0.7388** | **0.3881** | 0.0501 | 0.1118 |
| | GD | 0.7422 | 0.3879 | 0.0598 | 0.0931 | 0.7306 | 0.3924 | 0.0384 | 0.0741 |
| | DRO | 0.7316 | 0.4090 | **0.0136** | **0.0274** | 0.7230 | 0.4044 | 0.0340 | 0.0326 |
| | CMIA | 0.7160 | 0.4063 | 0.0144 | 0.0284 | 0.7304 | 0.4118 | **0.0058** | **0.0270** |
| | % | -3.72% | 3.85% | -79.89% | -71.96% | -1.14% | 6.08% | -88.49% | -75.82% |

Tables 1-4: Main results. The % rows denote the percentage change of CMIA compared to the vanilla models. ↑ means the larger the better and ↓ means the smaller the better.

To further compare all the competing methods' performance across the four different metrics, we employ radar charts visualization. We first apply the following linear transformation to the four metrics: $U_{AUC} = 0.5 + 2aAUC$, $U_{Brier} = 0.5 - aBrier$, $U_{dtpr} = -adTPR$, and $U_{dfpr} = -adFPR$, to get four new utility metrics (i.e., $U_{AUC}$, $U_{Brier}$, $U_{dtpr}$, $U_{dfpr}$). The purpose of the transformation is to rescale their values for more informative radar charts. In Figure 1, we show the radar charts on the COMPAS Multi dataset, where each radar dimension corresponds to one of the new utility metrics. A colored quadrilateral is formed for each competing method, and a larger area of quadrilateral indicates better overall performance.

As can be seen from Figure 1, our CMIA method leads to the largest area of quadrilateral across all the four scenarios, indicating the best overall performance in terms of both accuracy and fairness. Similar observations can be found in the radar charts for the other datasets; however, due to space constraints, they are not included here.

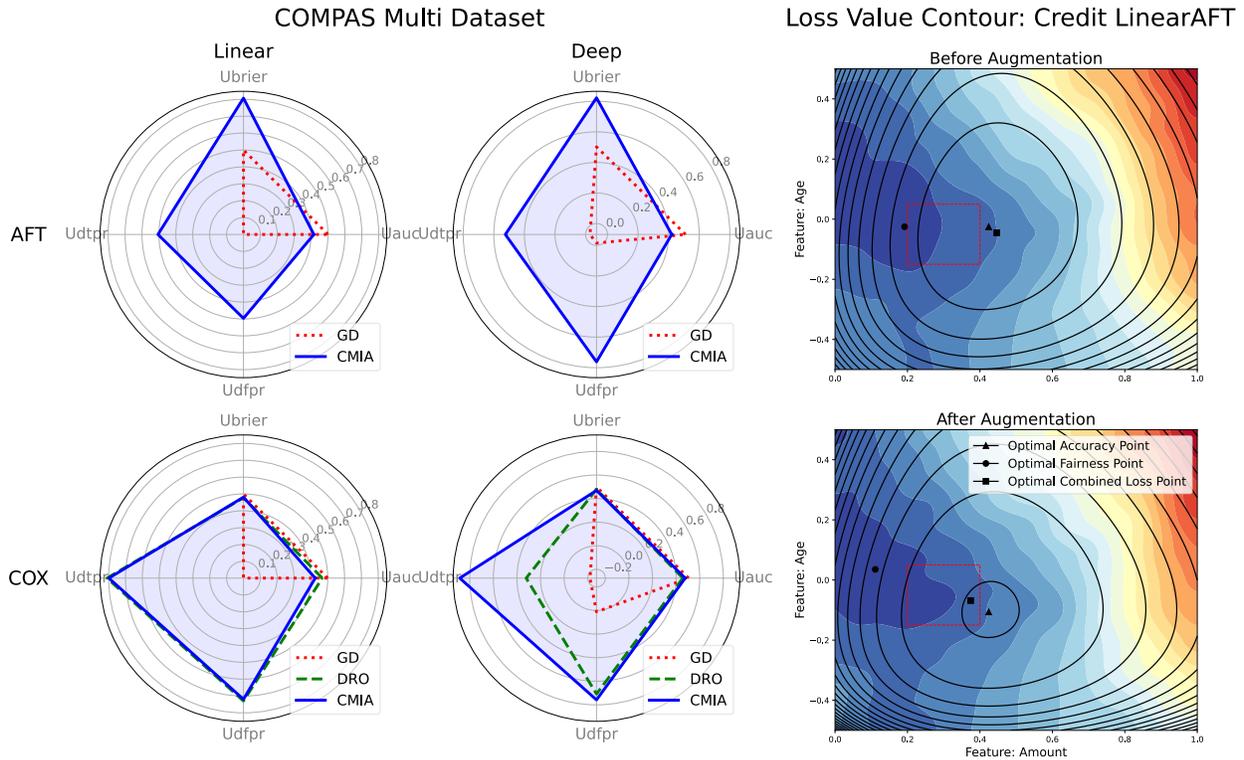

Figure 1: (on the left) Radar charts on the COMPAS Multi dataset.
Figure 2: (on the right) Contour plots of loss values before & after augmentation. The color-shaded area illustrates the contour of the fairness loss values, while the solid lines represent the contour of the accuracy loss values. The round dot indicates the coefficient combination that achieves the lowest fairness loss. The triangle dot marks the coefficient combination with the lowest accuracy loss. The rectangular dot shows the coefficient combination that results in the lowest combined loss, which is the sum of fairness loss and accuracy loss.

## 4.5. Ablation Study

Here we present an ablation study to validate the contribution of both components (i.e., the designed regularization term and the data augmentation module) to the performance of our CMIA approach. Specifically, we remove the regularization term from the CMIA approach and obtain a baseline method, denoted as CMIA-Reg, and removed the data augmentation and obtained another method, denoted as CMIA-Aug. We evaluate the performance of both CMIA-Reg and CMIA-Aug against the CMIA approach.

Due to space constraints, detailed results are not presented here. In summary, the CMIA-Aug method, which excludes the augmentation component, struggles to consistently address EO fairness. For instance, on the Credit dataset, CMIA-Aug produces a significantly higher adTPR compared to CMIA (0.139 vs. 0.0339). Conversely, the CMIA-Reg method, which removes the regularization component, exhibits unsatisfactory accuracy. Notably, on the FLC and Credit datasets, CMIA-Reg achieves an aAUC of only 50%, indicating predictions no better than random guessing. Our full CMIA approach, incorporating both the regularization and augmentation components, delivers the best overall performance.

## 4.6. Visualization: How Does Data Augmentation Help?

While the ablation study results quantitatively validate the effectiveness of the designed data augmentation module on balancing prediction accuracy and fairness, here we further employ visualization techniques to more intuitively interpret how the data augmentation improves the balance between accuracy and fairness. We consider the Credit dataset and the LinearAFT survival model scenario for presenting the interpretation.

Specifically, we first trained the vanilla model as described in Section 4.4. Next, we adjusted the values of the coefficients for the "amount" and "age" features, while keeping all other parameters fixed. For each combination of the two altered coefficients, we calculated the corresponding values of the foundational loss function $L(g_\theta, \Delta)$, the fairness regularization term $R_{EO}(g_\theta, \Delta)$, and the summation of both. We finally plotted these values against the various combinations of the altered coefficients to visualize the trade-off between accuracy and fairness.

As shown on the top of Figure 2, before data augmentation, there is a significant conflict between the fairness regularization and the foundational loss function. Specifically, the coefficients that minimize the fairness regularization (represented by the round dot) are located far from those that minimize the foundational loss function (represented by the triangle dot). Consequently, simply combining the two objectives in a joint learning framework fails to achieve fair results (represented by the rectangle dot), as the coefficients remain unwilling to move away from the optimal foundational loss point due to the large gap between the two objectives.

However, as illustrated in the dashed red rectangle, the contours shift after applying data augmentation. While the conflict between fairness regularization and the foundational loss function still exists, the coefficients become more inclined to move toward the fairness regularization minimum. This is because even a slight shift in the coefficients now leads to a substantial improvement in fairness, making it easier to reach the plateau with fairness values close to the optimal point. As a result, the rectangle dot moves closer to the round dot, leading to enhanced fairness without a significant compromise in accuracy.

## 5. CONCLUSIONS

This study makes several methodological contributions to the field. *First*, we proposed a novel fairness measurement, equalized odds (EO), for survival analysis. Unlike existing survival analysis fairness definitions, our proposed EO measurement emphasizes fairness across pre-defined time points, addressing a critical need in many real-world applications and filling an important gap in the literature. *Second*, to achieve the EO fairness, we developed a novel Conditional Mutual Information Augmentation (CMIA) approach. CMIA consists of two innovative methodological designs: a fairness regularization term based on conditional mutual information and a unique censored data augmentation technique. Both components are integrated into a joint learning objective, through which we infer the model parameters simultaneously. *Third*, one important merit of our CMIA approach is that it is model agnostic. We implemented and evaluated its performance by considering four different models of survival analysis, and the evaluation results demonstrate the superiority of our proposed CMIA approach.

Our developed CMIA approach could yield desired EO fairness in various survival analysis applications, leading to significant practical implications. *First*, in many high-stake domains such as healthcare and criminal justice, the EO fairness achieved by our CMIA approach could help practitioners make more equitable decisions such as fair medical interventions for healthcare. *Second*, the achieved EO fairness by our CMIA approach could eventually benefit individuals in those applications that rely on survival analysis-based prediction. For instance, it may prevent potential discrimination against minority groups in criminal justice when our CMIA approach is applied to predict future recidivisms.

The current study can be extended in several directions. *First*, it is worth investigating the performance of our CMIA approach in other application domains, as well as in other survival analysis scenarios that involve different $g_\theta(X)$ functions and objective functions. *Second*, another direction is to examine the long-term impacts of applying our fairness-aware approach in real-world applications, to further understand how it benefits different stakeholders.

# APPENDICES

## Appendix A: Proof of Equalized Odds Necessary and Sufficient Condition

$$\hat{Y}_t \perp Z | Y_t, \quad \forall t \in Q \iff P(\hat{Y}_t = 1 | Y_t = y, Z = c_i) = P(\hat{Y}_t = 1 | Y_t = y),$$

$$\forall t \in Q, \forall y \in \{0,1\}, \forall c_i \in C$$

$$\iff P(\hat{Y}_t = 1 | Y_t = y, Z = c_1) = \cdots = P(\hat{Y}_t = 1 | Y_t = y, Z = c_{|C|}),$$

$$\iff TPR_{c_1,t} = \cdots = TPR_{c_{|C|},t}; \; FPR_{c_1,t} = \cdots = FPR_{c_{|C|},t} \quad \forall t \in Q$$

## Appendix B: Proof of CMI Property

In this section, we prove the non-negativity of $\text{CMI}_t$ and its implication to Equalized Odds fairness when $\text{CMI}_t = 0$. By definition:

$$CMI_t := \sum_{\hat{Y}_t \in \{0,1\}} \sum_{Y_t \in \{0,1\}} \sum_{Z \in C} P_{\hat{Y}_t, Z, Y_t} \log \frac{P_{\hat{Y}_t | Z, Y_t}}{P_{\hat{Y}_t | Y_t}}, \quad \forall t \in Q$$

Applying Jensen's inequality, we have:

$$CMI_t \geq -\log \sum_{\hat{Y}_t \in \{0,1\}} \sum_{Y_t \in \{0,1\}} \sum_{Z \in C} P_{\hat{Y}_t, Z, Y_t} \frac{P_{\hat{Y}_t | Y_t}}{P_{\hat{Y}_t | Z, Y_t}}, \quad \forall t \in Q$$

$$= -\log \sum_{Y_t \in \{0,1\}} \sum_{Z \in C} P_{Z, Y_t} \sum_{\hat{Y}_t \in \{0,1\}} P_{\hat{Y}_t | Y_t}, \quad \forall t \in Q$$

$$= -\log \sum_{Y_t \in \{0,1\}} \sum_{Z \in C} P_{Z, Y_t} = -\log 1 = 0, \quad \forall t \in Q$$

Which concludes the proof for non-negativity. For proving its implication to Equalized Odds fairness when $\text{CMI}_t = 0$, we first observe that

$$CMI_t = \sum_{\widehat{Y}_t \in \{0,1\}} \sum_{Y_t \in \{0,1\}} \sum_{Z \in C} -P_{\widehat{Y}_t, Z, Y_t} \log \frac{P_{\widehat{Y}_t|Y_t}}{P_{\widehat{Y}_t|Z, Y_t}}, \quad \forall t \in Q$$

$$= \sum_{\widehat{Y}_t \in \{0,1\}} \sum_{Y_t \in \{0,1\}} \sum_{Z \in C} P_{\widehat{Y}_t, Z, Y_t} \left[ -\log \frac{P_{\widehat{Y}_t|Y_t}}{P_{\widehat{Y}_t|Z, Y_t}} + \frac{P_{\widehat{Y}_t|Y_t}}{P_{\widehat{Y}_t|Z, Y_t}} - 1 \right], \quad \forall t \in Q$$

Since the term $\left[ -\log \frac{P_{\widehat{Y}_t|Y_t}}{P_{\widehat{Y}_t|Z, Y_t}} + \frac{P_{\widehat{Y}_t|Y_t}}{P_{\widehat{Y}_t|Z, Y_t}} - 1 \right] \geq 0$ because $\log x \leq x - 1, \forall x \in \mathbb{R}$, we have

$$CMI_t = 0 \Rightarrow \log \frac{P_{\widehat{Y}_t|Y_t}}{P_{\widehat{Y}_t|Z, Y_t}} = \frac{P_{\widehat{Y}_t|Y_t}}{P_{\widehat{Y}_t|Z, Y_t}} - 1 \Rightarrow \frac{P_{\widehat{Y}_t|Y_t}}{P_{\widehat{Y}_t|Z, Y_t}} = 1 \Rightarrow \widehat{Y}_t \perp Z|Y_t, \quad \forall t \in Q$$

## Appendix C: Proof of CMI Approximation Efficiency

First, by applying the strong law of large number (SLLN), we can have:

$$\frac{1}{n_{Y_{i,t}Z_i}} \sum_{Y_{l,t}=Y_{i,t}, Z_l=Z_i} \phi_\tau \left( \epsilon_j + g_\theta(X_i) - g_\theta(X_l) \right) \xrightarrow{a.s.} \mathbb{E}_{X|Y_t=Y_{i,t}, Z=Z_i} \phi_\tau \left( \epsilon_j + g_\theta(X_i) - g_\theta(X) \right),$$

$$= \int_{X \in \mathbb{R}^p} \phi_\tau \left( \epsilon_j + g_\theta(X_i) - g_\theta(X) \right) P_{X|Y_t=Y_{i,t}, Z=Z_i} \, dX,$$

$$= \int_{X \in \mathbb{R}^p} P_{\epsilon_j + g_\theta(X_i)|X} \, P_{X|Y_t=Y_{i,t}, Z=Z_i} \, dX,$$

$$= P_{\epsilon_j + g_\theta(X_i)|Y_t=Y_{i,t}, Z=Z_i},$$

Similarly, we have:

$$\frac{1}{n_{Y_{i,t}}} \sum_{Y_{k,t}=Y_{i,t}} \phi_\tau \left( \epsilon_j + g_\theta(X_i) - g_\theta(X_k) \right) \xrightarrow{a.s.} \mathbb{E}_{X|Y_t=Y_{i,t}} \phi_\tau \left( \epsilon_j + g_\theta(X_i) - g_\theta(X) \right),$$

$$= \int_{X \in \mathbb{R}^p} \phi_\tau \left( \epsilon_j + g_\theta(X_i) - g_\theta(X) \right) P_{X|Y_t=Y_{i,t}} \, dX,$$

$$= \int_{X \in \mathbb{R}^p} P_{\epsilon_j + g_\theta(X_i)|X} \, P_{X|Y_t=Y_{i,t}} \, dX,$$

$$= P_{\epsilon_j + g_\theta(X_i)|Y_t=Y_{i,t}},$$

Therefore, when $m \longrightarrow \infty$, we can plug in the former asymptotic results and have:

$$\widehat{CMI}_{m,t} \coloneqq \frac{1}{n}\sum_{i=1}^{n}\frac{1}{m}\sum_{j=1}^{m} log \left[ \frac{\frac{1}{n_{Y_{i,t}Z_i}}\sum_{Y_{l,t}=Y_{i,t},Z_l=Z_i} \phi_\tau \left( \epsilon_j + g_\theta(X_i) - g_\theta(X_l) \right)}{\frac{1}{n_{Y_{i,t}}}\sum_{Y_{k,t}=Y_{i,t}} \phi_\tau \left( \epsilon_j + g_\theta(X_i) - g_\theta(X_k) \right)} \right], \quad \forall t \in Q$$

$$\xrightarrow{a.s.} \frac{1}{n}\sum_{i=1}^{n}\frac{1}{m}\sum_{j=1}^{m} log \left[ \frac{P_{\epsilon_j+g_\theta(X_i)|Y_t=Y_{i,t},Z=Z_i}}{P_{\epsilon_j+g_\theta(X_i)|Y_t=Y_{i,t}}} \right], \quad \forall t \in Q$$

Since $\epsilon_j \sim^{i.i.d} N(0,\tau)$, we can further apply SLLN to obtain:

$$\frac{1}{n}\sum_{i=1}^{n}\frac{1}{m}\sum_{j=1}^{m} log \left[ \frac{P_{\epsilon_j+g_\theta(X_i)|Y_t=Y_{i,t},Z=Z_i}}{P_{\epsilon_j+g_\theta(X_i)|Y_t=Y_{i,t}}} \right] \xrightarrow{a.s.} \mathbb{E}_{\epsilon,X,Y_t,Z} log \left[ \frac{P_{\epsilon+g_\theta(X)|Y_t,Z}}{P_{\epsilon+g_\theta(X)|Y_t}} \right],$$

Where the last term is the conditional mutual information $CMI(\epsilon + g_\theta(X), Z|Y_t)$. By the implication proved in the Appendix B, we have:

$$\widehat{CMI}_{m,t} \xrightarrow{a.s.} CMI(\epsilon + g_\theta(X), Z|Y_t) \geq 0,$$

which proves the asymptotic non-negativity of our approximation $\widehat{CMI}_{m,t}$. Apply the implication proved in Appendix B again, we have:

$$CMI(\epsilon + g_\theta(X), Z|Y_t) = 0 \Longrightarrow [\epsilon + g_\theta(X)] \perp Z|Y_t,$$

Since $\epsilon$ is drawn from an exogenous distribution independent from the data, we have:

$$[\epsilon + g_\theta(X)] \perp Z|Y_t \Longrightarrow g_\theta(X) \perp Z|Y_t,$$

Assuming the decision of $\hat{Y}_t$ solely depends on the predicted risk $g_\theta(X)$, and is expressed as an indicator function $\hat{Y}_t = I(g_\theta(X) \geq \alpha)$, where $\alpha \in (0,1)$ is a threshold (similar to the standard prediction protocol in classification tasks), we conclude the proof as follows:

$$g_\theta(X) \perp Z|Y_t \Longrightarrow \hat{Y}_t \perp Z|Y_t,$$